\documentclass{article}
\usepackage{spconf, amsmath, graphicx}
\usepackage{pifont, diagbox, multirow, float, wrapfig, booktabs, breqn}
\usepackage[dvipsnames]{xcolor}
\usepackage{soul}

\definecolor{MyPurple}{HTML}{990099}
\definecolor{MyGray}{HTML}{666666}
\newcommand{\cmark}{\text{\ding{51}}}
\newcommand{\xmark}{\text{\ding{55}}}

\title{FusionCount: Efficient Crowd Counting via Multiscale Feature Fusion}
\threeauthors
{Yiming Ma}
    {
    Warwick Mathematics Institute\\
    University of Warwick, UK
	}
{Victor Sanchez}
    {
    Department of Computer Science\\
    University of Warwick, UK
    }
{Tanaya Guha}
    {
    School of Computing Science\\
    University of Glasgow, UK
    }

\begin{document}
\date{}
\ninept
\maketitle
\begin{abstract}
State-of-the-art crowd counting models follow an encoder-decoder approach. Images are first processed by the encoder to extract features. Then, to account for perspective distortion, the highest-level feature map is fed to extra components to extract multiscale features, which are the input to the decoder to generate crowd densities. However, in these methods, features extracted at earlier stages during encoding are underutilised, and the multiscale modules can only capture a limited range of receptive fields, albeit with considerable computational cost. This paper proposes a novel crowd counting architecture (\emph{FusionCount}), which exploits the adaptive fusion of a large majority of encoded features instead of relying on additional extraction components to obtain multiscale features. Thus, it can cover a more extensive scope of receptive field sizes and lower the computational cost. We also introduce a new channel reduction block, which can extract saliency information during decoding and further enhance the model's performance. Experiments on two benchmark databases demonstrate that our model achieves state-of-the-art results with reduced computational complexity.
\end{abstract}

\begin{keywords}
Crowd density estimation, multiscale feature fusion, efficient crowd counting
\end{keywords}

\section{Introduction}
\label{sec:intro}

Crowd counting aims to automatically estimate the number of individuals present in a scene from an image or video. It can be applied in numerous areas, such as traffic control \cite{6514618}, biological studies \cite{lu2017tasselnet}, and recently,  social distancing monitoring \cite{9562868}.

Over the years, models for crowd counting have evolved from using classical regression models, such as random forests \cite{lempitsky2010learning} and Gaussian processes \cite{5459191}, to high-performing convolutional neural networks (CNNs) \cite{MCNN, CSRNet, CAN, BayesianLoss}. These deep nets usually adopt an encoder-decoder approach: First, an image is fed to the encoder to learn data representations (feature maps). The decoder then exploits the highest-level representation (the output from the encoder's last layer) to generate the density map, which is the distribution of the crowd. Since the convolutional and pooling blocks of VGG networks \cite{VGG}, where each layer exploits kernels of a fixed size, constitute most encoders, the size of receptive fields remains constant across the last encoded feature map. Thus, this representation can only handle images where crowds are of similar scales. However, people are usually depicted in various sizes because of the camera perspective, and as a result, the encoded feature should also have different receptive field sizes to model the scale variation. Multi-column structures \cite{MCNN, SANet} have been proposed to solve this problem. However, it has been recently shown that features from each column are almost identical, and training deep models of this type can be very unproductive \cite{CSRNet}. Therefore, to solve this scale issue, state-of-the-art methods \cite{CAN, M-SFANet, SMANet} employ a multiscale module that further processes the encoded representation and generates a feature map with different receptive field sizes. However, such a strategy ignores that those feature maps extracted by shallower encoding layers already provide information about different scales, and leveraging extra components makes the overall model more computationally expensive. 

Hence, our contribution in this paper is a novel multiscale mechanism that addresses the scale issue by leveraging the majority of features generated from the encoder to avoid extra feature-extraction modules and keep the computational cost low. This design incorporates a comprehensive range of receptive field sizes (6 to 192), covering 
almost all possible scales a person can depict in a crowd image. Experiments on two benchmark databases (ShanghaiTech A \& B \cite{MCNN}) demonstrate that our model can achieve state-of-the-art or comparable results with significantly fewer floating-point operations.

\section{Related Work}
\label{sec:related work}

Early crowd counting approaches \cite{983420, 4761705, 5206621} are based on object detectors, while later works tend to avoid them because of their sensitivity to occlusion and the enormous efforts required to annotate bounding boxes. Some of these non-detection-based methods \cite{lempitsky2010learning, 5459191, chen2012feature, 6619163} treat crowd counting as a regression problem: they learn low-level feature representations first, from which the total count is then directly regressed. The training losses of these approaches depend only on the ground-truth count (a scalar) and do not consider crowd density distribution, hence suffering poor generalisation. Thus, these models have soon been superseded by algorithms \cite{CP-CNN, Hydra-CNN, CAN, M-SFANet} that instead predict crowd densities, and these density-based methods primarily rely on CNNs.

\begin{figure*}[t]
    \centering
    \includegraphics[width=\textwidth]{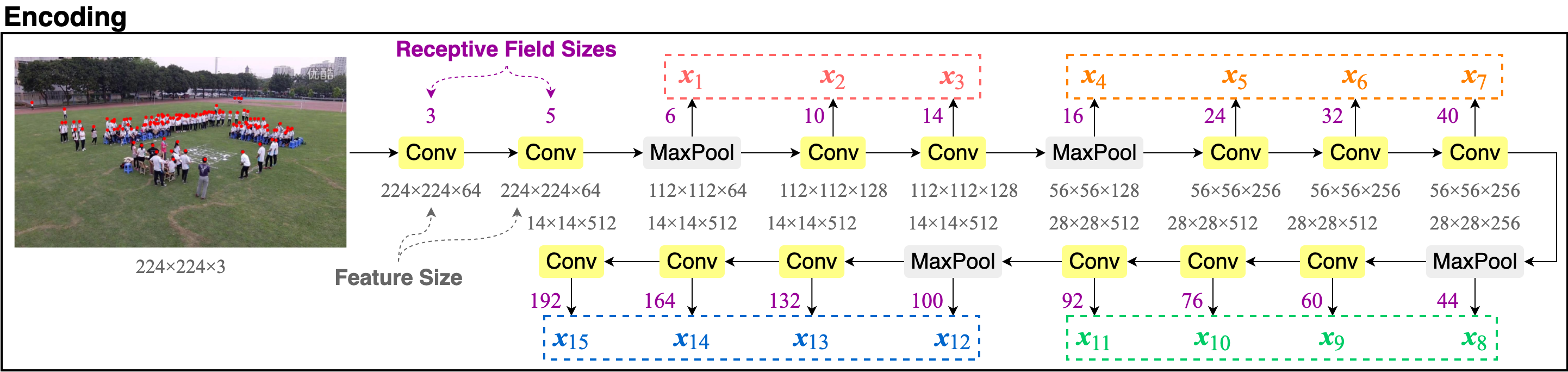}
    \caption{The encoder of our proposed model FusionCount: Only the first 17 layers of the original VGG-16 are leveraged, and feature maps are collected starting from the third layer. Numbers in \textcolor{MyPurple}{purple} are features' receptive field sizes and those tuples ($h \times w \times c$) in \textcolor{MyGray}{gray} indicate their sizes, assuming the input image has the size of $224\times224\times3$. Features with the same spatial resolution are grouped together for the first-phase fusion.}
    \label{fig:encoder}
\end{figure*}

Since 3D spatial locations have to be projected onto a 2D space while an RGB crowd image is captured from the real world, people can be depicted in different sizes due to perspective distortion, and the variation in scales can severely affect density estimation. To solve this problem, in \cite{Geometric}, extra geometric information is exploited to adapt their model to different scenes, but this information is not always provided. Therefore, later methods tend to learn the scales of crowds implicitly. For example, Hydra-CNN \cite{Hydra-CNN} divides an image into a pyramid of patches, each representing a different scale and fed to a separate encoder head. Then, all encoded features are concatenated without further processing and utilised to generate the density map. This approach neglects the fact that scales vary continuously across the whole image. CAN \cite{CAN} is then proposed to address this issue. The whole model involves only one encoder, so a spatial pyramid pooling module \cite{SpatialPyramidPooling} is leveraged to make it scale-aware. Then, features are averaged according to learnable weights to ensure that receptive field sizes of the fused feature changes smoothly. However, this architecture is less efficient since it does not exploit low-level features extracted during encoding. These representations, along with the final high-level feature map, can provide information about different scales because they have disparate receptive field sizes. Also, the multiscale module in \cite{CAN} uses only four filter sizes, thereby covering a limite range of scales.

\section{Our Model}
\label{sec:method}

This section exhaustively describes our proposed crowd counting model, \emph{FusionCount}. It extensively leverages representations learned during encoding to compute \emph{first-phase} multiscale features, and its decoder further fuses these scale-aware features to generate the density map.

\subsection{Encoding}
\label{subsec:encoding}

Following common practices \cite{CSRNet, CAN} in the field, VGG-16 \cite{VGG} is employed as our model's encoder. The last max-pooling and fully connected layers are removed since they are accountable for class prediction. Therefore, the encoder comprises 13 convolutional layers and four max-pooling layers. Since the two feature maps extracted before the first max-pooling operation are not sufficiently informative, only features learned afterwards are preserved and fused. Thus, as shown in Fig.~\ref{fig:encoder}, the encoder outputs 15 feature maps in total, denoted by $\boldsymbol{x}_j$, $j = 1, \, 2, \, \cdots, \, 15$. These feature maps are divided into four groups according to their heights and widths: $\boldsymbol{x}_{1}$ -- $\boldsymbol{x}_{3}$, $\boldsymbol{x}_{4}$ -- $\boldsymbol{x}_{7}$, $\boldsymbol{x}_{8}$ -- $\boldsymbol{x}_{11}$, and $\boldsymbol{x}_{12}$ -- $\boldsymbol{x}_{15}$.

\subsection{Feature Fusion}
\label{ssec:feature_fusion}

\begin{figure}[t]
    \centering
    \includegraphics[width=0.48\textwidth]{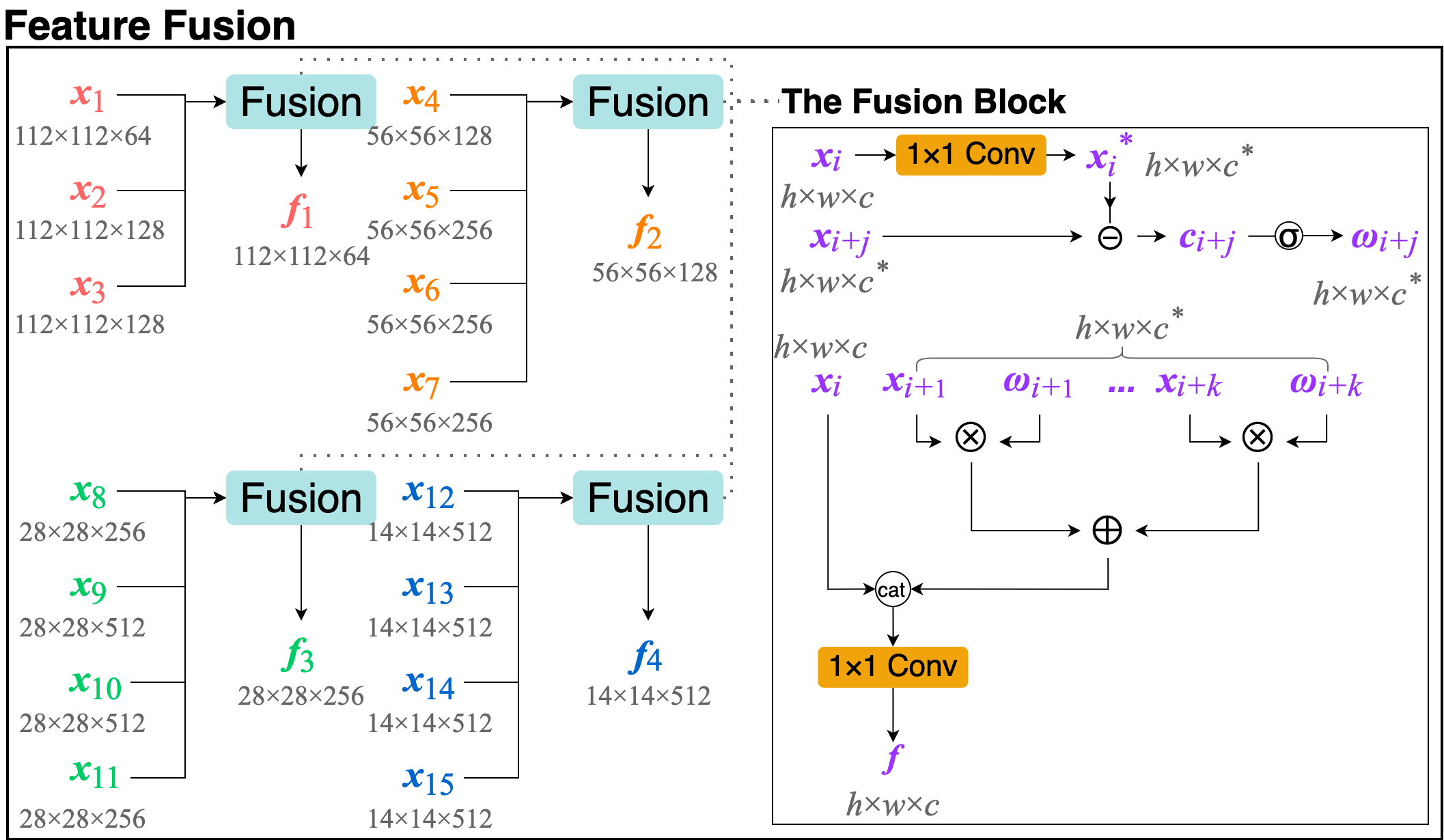}
    \caption{The feature fusion modules of FusionCount: In each group, weights are computed from contrast features. Then features from convolutional layers are averaged by using these weights and subsequently concatenated with the feature map from the pooling layer.}
    \label{fig:feature_fusion}
\end{figure}

Suppose $\boldsymbol{x}_i, \, \boldsymbol{x}_{i+1}, \, \cdots , \,  \boldsymbol{x}_{i+k}$ is a group of feature maps of interest, where $\boldsymbol{x}_i$ is generated by a max-pooling layer and has $c$ channels, while the rest are are produced by convolutional layers and have $c^*$ channels. To assemble a scale-aware representation from them, we adopt a similar strategy as the one proposed in \cite{CAN}, and our method is illustrated in Fig.~\ref{fig:feature_fusion}. Given that scales change continuously accross an image, the receptive field size of the output multiscale feature, which models the scales, should also be spatially continuous. To accomplish this aim, we fuse these features via weighted averaging, where weight maps are learned from each feature's spatial importance.

Firstly, since $\boldsymbol{x}_i$ and $\boldsymbol{x}_{i+j}$ ($j \in \{1, \cdots, k\}$) may not have the same number of channels, we use point-wise convolution to expand $\boldsymbol{x}_i$, and denote the output by $\boldsymbol{x}_i^*$. Notice that since $\boldsymbol{x}_i$ is the output of a max-pooling layer and the derivation of $\boldsymbol{x}_i^*$ does not involve any integration of neighbouring spatial information, $\boldsymbol{x}_i^*$ is less influenced by background noise. With such an advantage, all other representations are compared against it. Namely, following \cite{CAN}, we propose a similar concept to \emph{contrast features}:
\begin{equation}
    \label{eqn1}
    \boldsymbol{c}_{j} = \boldsymbol{x}_{i+j} - \boldsymbol{x}_i^*,
\end{equation}
with $j = 1, \, \cdots, k$. Since $\boldsymbol{x}_i^*$ and $\boldsymbol{x}_{i+j}$ ($j=1, \, \cdots, \, k$) have different receptive field sizes, the contrast features $\boldsymbol{c}_j$ incorporate disparities between any spatial location and its neighbouring pixels. Thus, they can facilitate learning the boundary of each person, thereby determining each receptive field size's relative importance. We then compute the weights $\boldsymbol{\omega}_{j}$ as follows:
\begin{equation}
    \boldsymbol{\omega}_{j} = \sigma (\boldsymbol{c}_{j}),
\end{equation}
where $\sigma$ denotes the sigmoid function.\footnote{Based on experiements (see Section \ref{ssec:contrast_features}), the performance of the overall model drops if contrast features are not utilised to compute weights.} Using these weights, $\boldsymbol{\omega}_{i+1}, \cdots \boldsymbol{\omega}_{i+k}$,  we then combine $\{ \boldsymbol{x}_i, \, \boldsymbol{x}_{i+1}, \, \cdots, \, \boldsymbol{x}_{i+k} \}$ adaptively and concatenate the averaged feature map with $\boldsymbol{x}_i$. Finally, we add a bottleneck layer to reduce computation. This process can be expressed as:
\begin{equation}
    \boldsymbol{f} = \text{Conv} \left(\left[ \boldsymbol{x}_i \, \middle| \, \sum_{j=1}^k \boldsymbol{\omega}_j \odot  \boldsymbol{x}_{i + j} \right] \right),
    \label{multiscale}
\end{equation}
where $\boldsymbol{f}$, Conv, ``$|$'' and ``$\odot$'' denote the fused feature, the bottleneck layer, channel-wise concatenation, and element-wise product, respectively.

\subsection{Decoding}
\label{ssec:decoding}

\begin{figure}[t]
    \centering
    \includegraphics[width=0.48\textwidth]{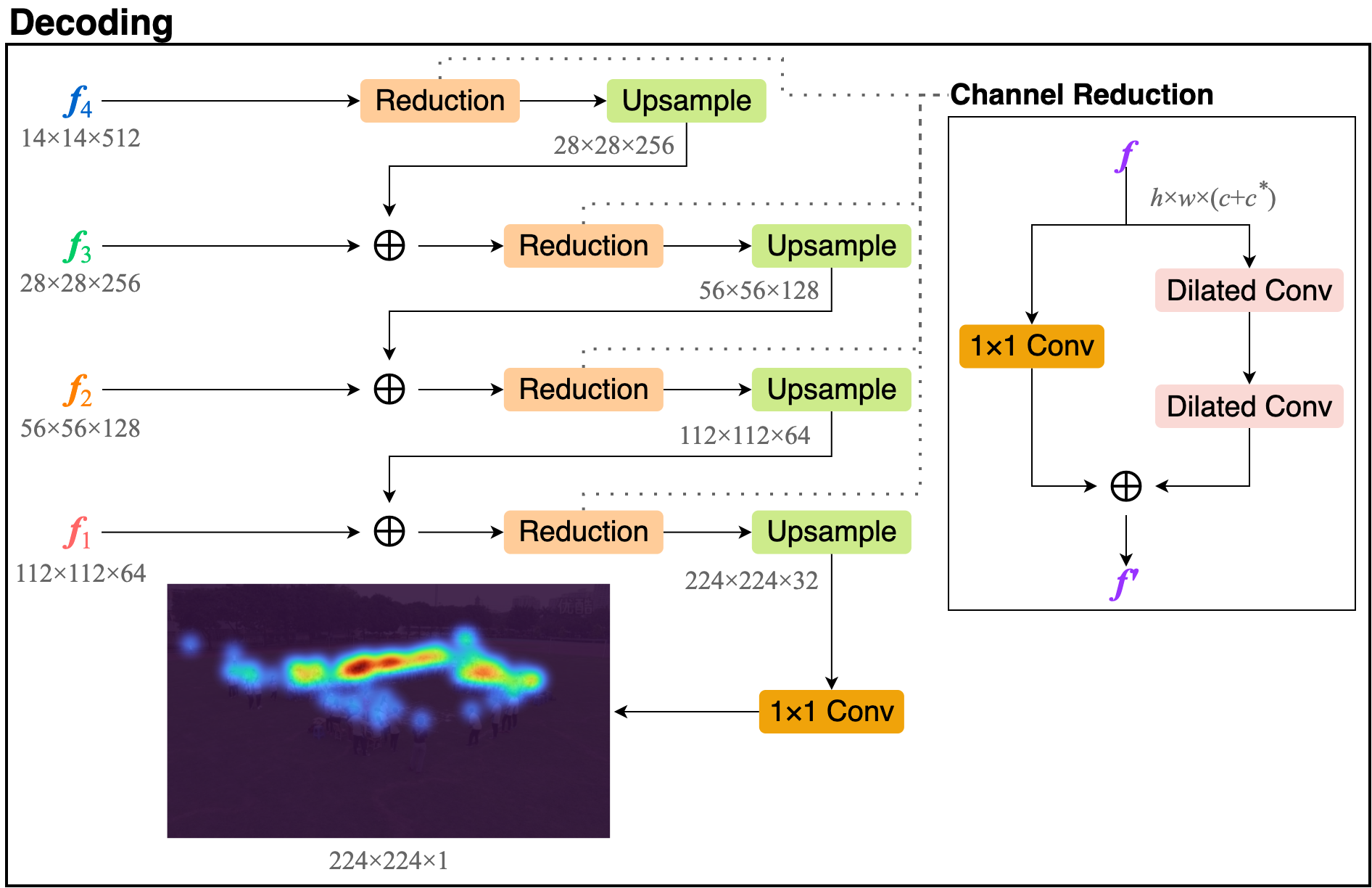}
    \caption{The decoding process of FusionCount: starting from $\boldsymbol{f}_4$, the proposed channel reduction module first decreases its number of channels. The result is then upsampled and fused with another firs-phase multiscale feature $\boldsymbol{f}_3$.}
    \label{fig:decoder}
\end{figure}

As shown in Fig.~\ref{fig:feature_fusion}, we denote the first-phase multiscale features generated from the four groups via \eqref{multiscale} as $\boldsymbol{f}_1$, $\boldsymbol{f}_2$, $\boldsymbol{f}_3$ and $\boldsymbol{f}_4$. These features are further fused in the reverse order (see Fig.~\ref{fig:decoder}). To combine $\boldsymbol{f}_4$ and $\boldsymbol{f}_3$ via addition, we first transform the shape of $\boldsymbol{f}_4$. Specifically, the number of channels of $\boldsymbol{f}_4$ is lowered, while its spatial size is expanded. Traditionally, a single point-wise convolutional kernel is used to reduce the nubmer of channels. However, inspired by dilated convolution, which has been shown to have the ability to extract deeper salient information while maintaining the spatial resolution \cite{CSRNet}, we propose a new channel reduction module. Our two-stream module comprises a dilated convolution block and one bottleneck layer, and the two columns are connected via addition. \footnote{The effectiveness of the combination of the dilated convolution block and the point-wise convolution is verified in Section \ref{ssec:ablation_study}.} To match the spatial resolution of $\boldsymbol{f}_3$, we use bilinear interpolation to double the size of  $\boldsymbol{f}_4$ after channel redcution. Then these two features can be fused by summation. The new fused feature is then combined with $\boldsymbol{f}_2$ following a similar pattern --- we first modify its dimensions via the proposed channel reduction module and bilinear interpolation, and then add it to  $\boldsymbol{f}_2$. This process is followed iteratively until all features are fused. Finally, we feed the final fused feature to the output layer to generate the estimation.

\section{Experiments}
\label{sec:experiments}

\subsection{Datasets}
\label{ssec:datasets}

We use the ShanghaiTech A \& B datasets \cite{MCNN} for model evaluation and comparison. In ShanghaiTech A, there are 482 crowd images collected from the Internet. Three hundred of them constitute the training set, and the rest comprise the test set. Scenes in this dataset are highly congested, with an average count of about 501. Also, since images have different resolutions (height and width values range from 182 to 1024), training models on this dataset can be tricky. The size of ShanghaiTech B is larger (716 instances in total; 400 for training and 316 for testing), with an average count of approximately 123. Images from this dataset are taken from a surveillance view in a shopping street and are therefore less crowded. Also, considering that these images have a fixed resolution ($768\times1024$), this dataset is more suitable for real-world applications e.g., public safety monitoring.

\subsection{Experiment Settings}
\label{ssec:settings}

During recent years, loss functions based on probability theory, such as Bayes' theorem and Wasserstein distance, have been shown to help models achieve stronger generalisation capabilities and have thus gained increased popularity. We use the DM-Count loss \cite{DMCount} to supervise the training of FusionCount. An Adam optimiser \cite{Adam} with a learning rate 1e-5 and batch size two is leveraged for optimisation. In order to make fair comparisons with other approaches, we use the default data splits. Given that some images have intractably large sizes, from each input image, two patches with a size of $384\times512$ are cropped and used for training. Our model and its training are implemented in the PyTorch \cite{PyTorch} 1.10 framework, and the platform for training is a server with an NVIDIA RTX 3090 GPU and Ubuntu 20.04 LTS OS.

\subsection{Results}
\label{ssec:results}

\begin{table}[t]
\centering \scriptsize
\renewcommand*{\arraystretch}{1.4}
\caption{Comparision of our model FusionCount with state-of-the-art models of similar sizes. The best and the second best results are indicated in \textbf{bold} and  \underline{underlined} typefaces, respectively.}
\begin{tabular}{ l | c | c | c | c | c }
\toprule
\multirow{2}{*}{\bf Model} & \multirow{2}{*}{\bf Mult-Adds} & \multicolumn{2}{c|}{\bf SH A} & \multicolumn{2}{c}{\bf SH B} \\ \cline{3-6}
& & MAE & RMSE & MAE & RMSE \\
\midrule
CSRNet \cite{CSRNet} & 856.99 G & 68.2 & 115.0 & 10.6 & 16.0 \\
CAN \cite{CAN} & 908.05 G & 62.3 & \underline{100.0} & 7.8 & \underline{12.2} \\
BL \cite{BayesianLoss} & \underline{853.70 G} & 62.8 & 101.8 & 7.7 & 12.7 \\
DM-Count \cite{DMCount} & \underline{853.70 G} & \textbf{59.7} & \textbf{95.7} & \underline{7.4} & \textbf{11.8} \\
\bf FusionCount (ours) & \textbf{815.00 G} & \underline{62.2} & 101.2 & \textbf{6.9} & \textbf{11.8} \\
\bottomrule
\end{tabular}
\label{table:2}
\end{table}

Following prior works \cite{CSRNet, CAN, DMCount}, the mean absolute error (MAE) and the root mean squared error (RMSE) of total counts are employed as evaluation metrics.
FusionCount is compared with state-of-the-art models that have a similar computational complexity (quantified by the number of multiplications and additions involved in the inference on a $1080\times1920$ RGB image). In particular, CSRNet \cite{CSRNet}, CAN \cite{CAN}, BL \cite{BayesianLoss} and DM-Count \cite{DMCount}, all of which employ VGGs \cite{VGG} as encoders, are encompassed for comparison. CSRNet uses dilated convolution in decoding to extract saliency. In addition to this characteristic, CAN also includes a spatial pyramid pooling block to generate multiscale features. BL and DM-Count have no innovations in terms of model architectures, and their contributions are particularly novel loss functions based on probability theory. Table \ref{table:2} shows detailed performance comparisons. For the case of ShanghaiTech A, partly due to the difficulty in training, FusionCount's results are not the best but still comparable: it achieves the second-lowest mean absolute counting error. On ShanghaiTech B, FusionCount outperforms the other evaluated models under both metrics. These results are remarkable, especially considering our model's low computational complexity.

\begin{figure}[H]
\centerline{GT: 1603; Pred: 1634.79; RE: 1.98\%} \medskip
\vspace{-0.5em}
\begin{minipage}[t]{0.49\linewidth}
    \centering
    \centerline{\includegraphics[width=4.0cm]{images/sha_img_116_orig.jpg}}
\end{minipage}
\hfill
\begin{minipage}[t]{0.49\linewidth}
    \centering
    \centerline{\includegraphics[width=4.0cm]{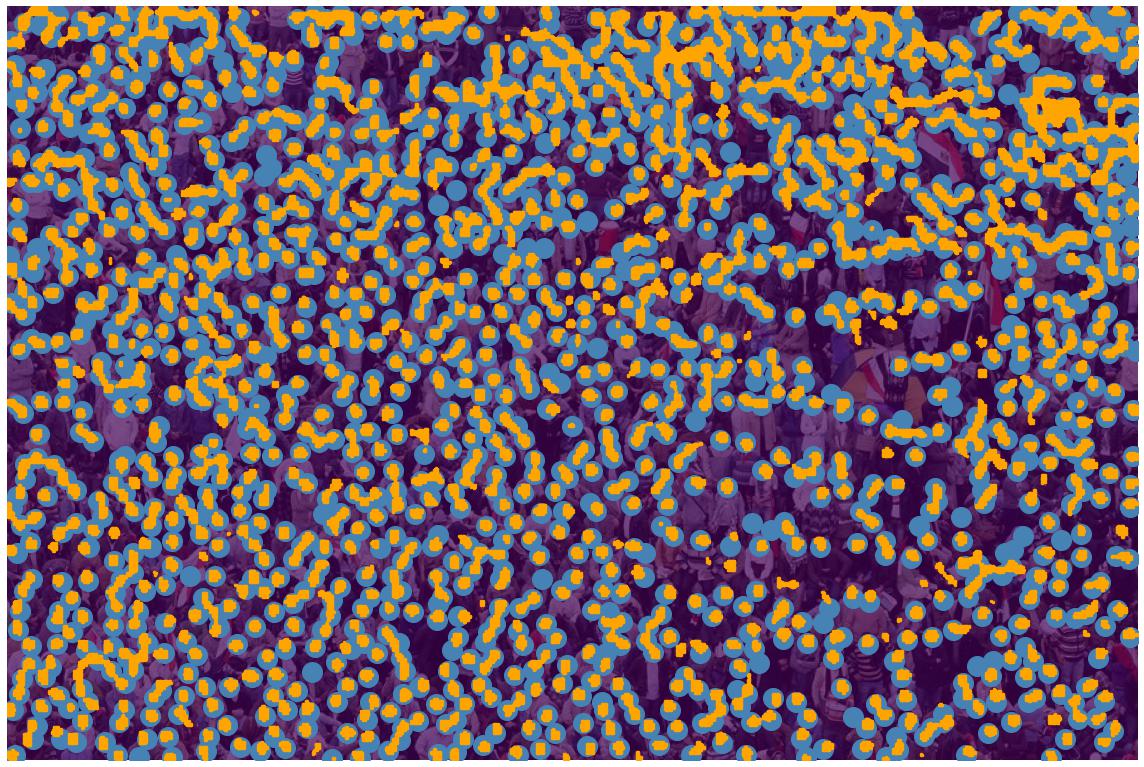}}
\end{minipage}
\centerline{GT: 521; Pred: 525.97; RE: 0.95\%} \medskip
\begin{minipage}[t]{0.49\linewidth}
    \centering
    \centerline{\includegraphics[width=4.0cm]{images/sha_img_96_orig.jpg}}
\end{minipage}
\hfill
\begin{minipage}[t]{0.49\linewidth}
    \centering
    \centerline{\includegraphics[width=4.0cm]{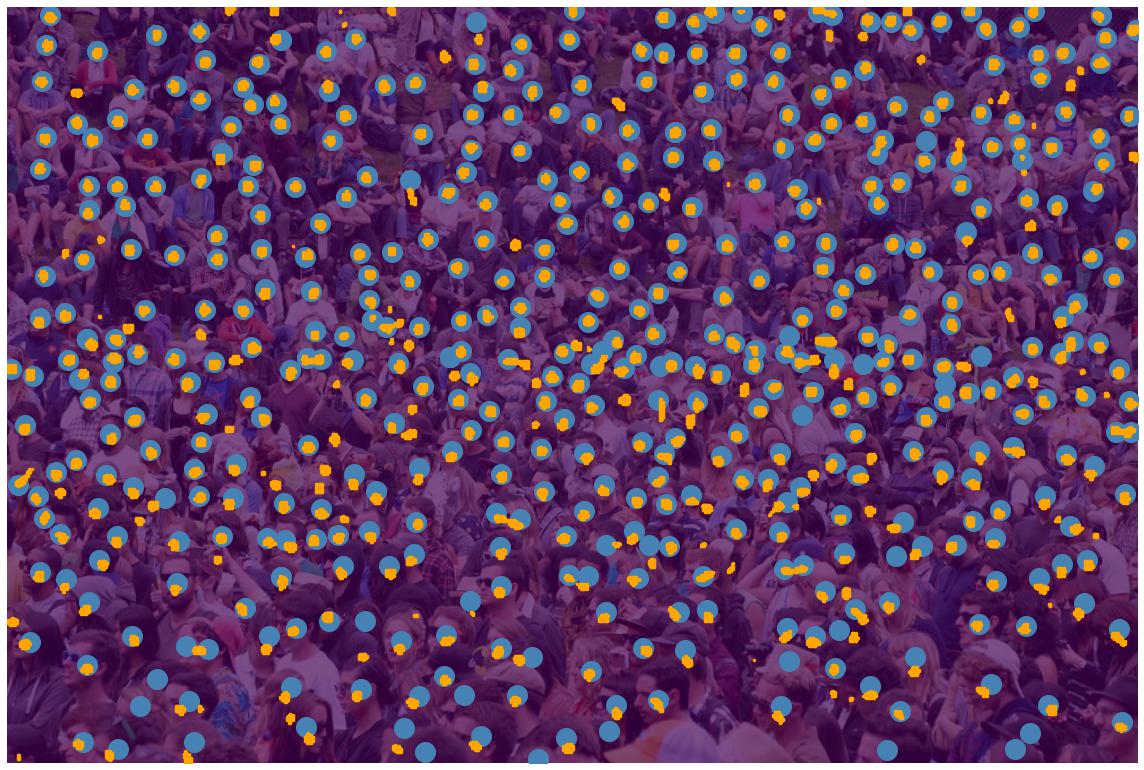}}
\end{minipage}
\centerline{GT: 300; Pred: 302.10; RE: 0.70\%} \medskip
\vspace{-0.5em}
\begin{minipage}[t]{0.49\linewidth}
    \centering
    \centerline{\includegraphics[width=4.0cm]{images/shb_img_21_orig.jpg}}
\end{minipage}
\hfill
\begin{minipage}[t]{0.49\linewidth}
    \centering
    \centerline{\includegraphics[width=4.0cm]{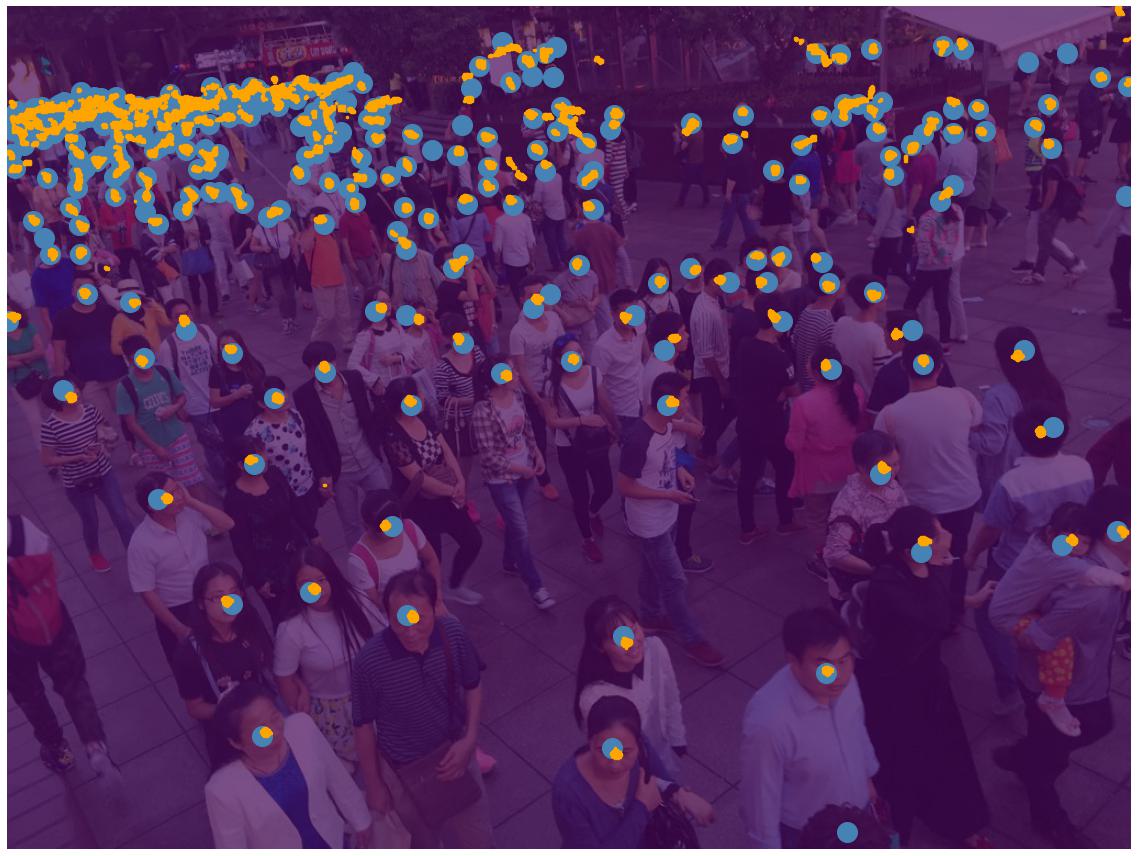}}
\end{minipage}
\centerline{GT: 176; Pred: 176.08; RE: 0.05\%} \medskip
\begin{minipage}[t]{0.49\linewidth}
    \centering
    \centerline{\includegraphics[width=4.0cm]{images/shb_img_30_orig.jpg}}
\end{minipage}
\hfill
\begin{minipage}[t]{0.49\linewidth}
    \centering
    \centerline{\includegraphics[width=4.0cm]{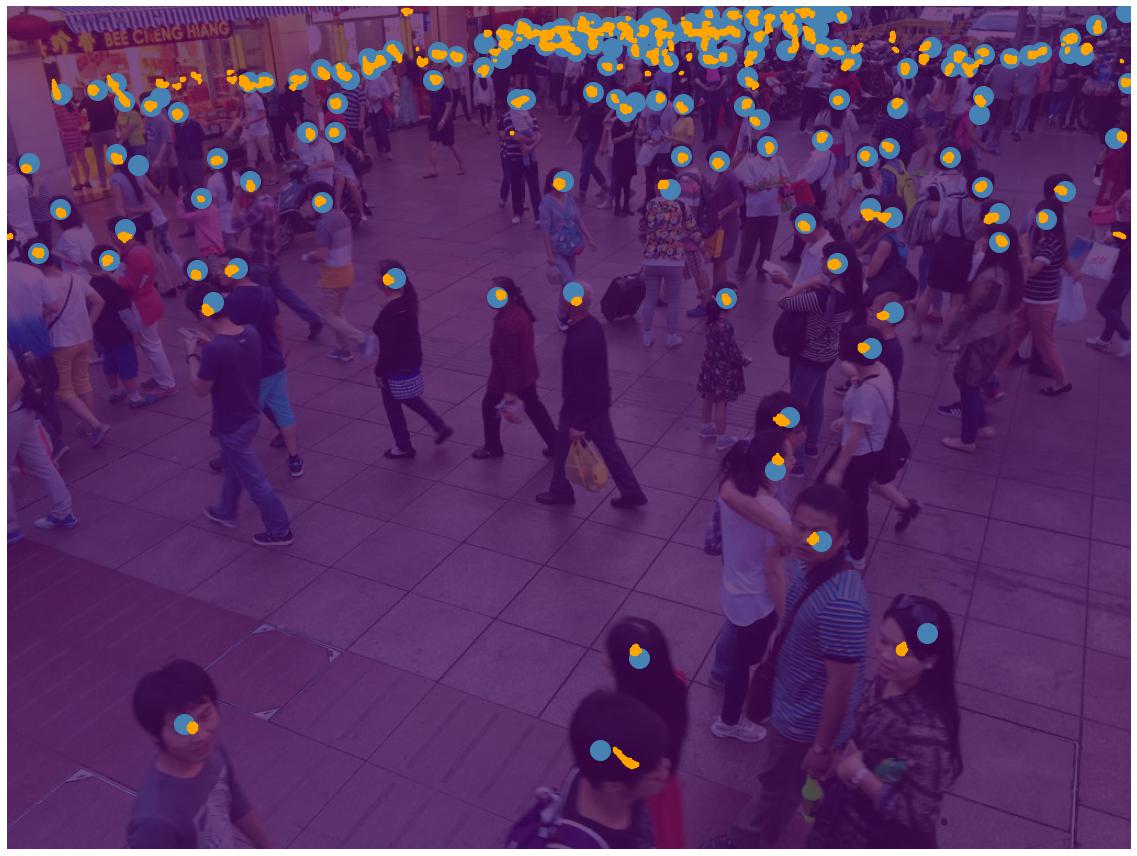}}
\end{minipage}
\caption{Qualitative results of FusionCount: Four ground-truth density maps are visualised along with their estimations by our model FusionCount. The top two images are from ShanghaiTech A and the bottom two are from ShanghaiTech B. Ground-truth annotations are marked by \textcolor{NavyBlue}{blue} dots, and the \textcolor{YellowOrange}{orange} shades represent the estimated densities. We also report `GT' i.e., ground-truth counts of people in the images, `Pred' i.e., the predicted counts. `RE' denotes the correpsonding relateive error.}
\label{fig:shab_vis}
\end{figure}

Four instances from the test sets of ShanghaiTech A \& B are depicted in Fig.~\ref{fig:shab_vis}. The left column shows the original input images, and the right column illustrates the ground-truth and the predicted density maps, which are indicated by blue dots and orange shades, respectively. The example in the first row proves that our model can still work well in highly crowded cases. Although in this $600 \times 900$ image, there are over 1,600 people, our model still achieves an accurate prediction with a relatively small error (1.98\%). Other rows demonstrate that our model is capable of effectively dealing with scale changes. In these images, crowds in the lower parts of the scene have larger scales, while those in the upper part of the scene have the smallest scales. Our model can make correct predictions for both cases with negligible errors (0.95\%, 0.70\% and 0.05\%, respectively). Thus, all these four instances confirm our model's strong counting ability.

\subsection{Effectiveness of Contrast Features}
\label{ssec:contrast_features}

To confirm that the contrast features can boost the model's performance, we create a variant for our model, whose only difference from FusionCount is the feature fusion strategy. In this variant, weights are directly generated from encoded features $\boldsymbol{x}_{i+j}$ instead of contrast features $\boldsymbol{c}_{i+j}$. We train this variant on ShanghaiTech B \cite{MCNN} with the same setting. The MAE and RMSE of this variant are 7.6 and 12.9, respectively, which are larger than those of FusionCount (6.9 and 11.8).

\subsection{Ablation Study}
\label{ssec:ablation_study}

This section proves the effectiveness of the channel reduction module proposed in Section \ref{ssec:decoding}. Experiments are conducted on ShanghaiTech B \cite{MCNN}. The results, which are shown in Table \ref{table:3}, demonstrate that both the point-wise convolutional layer and the two chained dilated convolutional layers are indispensable. Theoretically, they work in a complementary way in reducing the number of channels and extract saliency.

\begin{table}[htbp]
\centering 
\caption{Effects of point-wise convolution and the number of dilated convolutional layers in the channel reduction module.}
\vspace{1mm}
\renewcommand*{\arraystretch}{1.4}
\begin{tabular}{ c | c | c | c | c | c }
\toprule
\multirow{2}{*}{Point-wise Conv} & \multicolumn{3}{c|}{Dilated Conv No.} & \multirow{2}{*}{MSE} & \multirow{2}{*}{RMSE} \\
\cline{2-4}
              & 2      & 1      & 0      &              &  \\
\cline{1-6}
\cmark        & \cmark &        &        & \textbf{6.9} & \textbf{11.8} \\
\cmark        &        & \cmark &        & 7.6          & 13.0 \\
\cmark        &        &        & \cmark & 8.8          & 15.0 \\
\xmark        & \cmark &        &        & 9.5          & 15.9 \\
\bottomrule
\end{tabular}
\label{table:3}
\end{table}

\section{Conclusion}
\label{sec:conclusion}

In this paper, we proposed a new crowd counting architecture, FusionCount, which smartly utilises a large majority of features generated during the encoding process to handle perspective distortion. Unlike existing approaches, FusionCount avoids further extraction of multiscale features, thereby significantly reducing overall computation. To this end, We have also improved an existing multiscale fusion mechanism and devised a novel channel reduction block. Experiments on the ShanghaiTech databases demonstrated that FusionCount can outperform relevant state-of-the-art approaches of similar computational complexity. As part of our future work, we are working on accounting for any contextual information in the features fused at the decoding process. Such information can help to more effectively deal with scale changes, as the way the first-stage fusion of encoded features does.


\bibliographystyle{IEEEbib}
\bibliography{refs}

\begin{thebibliography}{10}

\bibitem{6514618}
Thomas Moranduzzo and Farid Melgani,
\newblock ``Automatic car counting method for unmanned aerial vehicle images,''
\newblock {\em IEEE Transactions on Geoscience and Remote Sensing}, vol. 52,
  no. 3, pp. 1635--1647, 2014.

\bibitem{lu2017tasselnet}
Hao Lu, Zhiguo Cao, Yang Xiao, Bohan Zhuang, and Chunhua Shen,
\newblock ``Tasselnet: counting maize tassels in the wild via local counts
  regression network,''
\newblock {\em Plant methods}, vol. 13, no. 1, pp. 1--17, 2017.

\bibitem{9562868}
Immanuel Jose~C. Valencia, Elmer~P. Dadios, Alexis~M. Fillone, John Carlo~V.
  Puno, Renann~G. Baldovino, and Robert Kerwin~C. Billones,
\newblock ``Vision-based crowd counting and social distancing monitoring using
  tiny-yolov4 and deepsort,''
\newblock in {\em 2021 IEEE International Smart Cities Conference (ISC2)},
  2021, pp. 1--7.

\bibitem{lempitsky2010learning}
Victor Lempitsky and Andrew Zisserman,
\newblock ``Learning to count objects in images,''
\newblock in {\em Advances in Neural Information Processing Systems},
  J.~Lafferty, C.~Williams, J.~Shawe-Taylor, R.~Zemel, and A.~Culotta, Eds.
  2010, vol.~23, Curran Associates, Inc.

\bibitem{5459191}
Antoni~B. Chan and Nuno Vasconcelos,
\newblock ``Bayesian poisson regression for crowd counting,''
\newblock in {\em 2009 IEEE 12th International Conference on Computer Vision},
  2009, pp. 545--551.

\bibitem{MCNN}
Yingying Zhang, Desen Zhou, Siqin Chen, Shenghua Gao, and Yi~Ma,
\newblock ``Single-image crowd counting via multi-column convolutional neural
  network,''
\newblock in {\em 2016 IEEE Conference on Computer Vision and Pattern
  Recognition (CVPR)}, 2016, pp. 589--597.

\bibitem{CSRNet}
Yuhong Li, Xiaofan Zhang, and Deming Chen,
\newblock ``Csrnet: Dilated convolutional neural networks for understanding the
  highly congested scenes,''
\newblock in {\em 2018 IEEE/CVF Conference on Computer Vision and Pattern
  Recognition}, 2018, pp. 1091--1100.

\bibitem{CAN}
Weizhe Liu, Mathieu Salzmann, and Pascal Fua,
\newblock ``Context-aware crowd counting,''
\newblock in {\em 2019 IEEE/CVF Conference on Computer Vision and Pattern
  Recognition (CVPR)}, 2019, pp. 5094--5103.

\bibitem{BayesianLoss}
Zhiheng Ma, Xing Wei, Xiaopeng Hong, and Yihong Gong,
\newblock ``Bayesian loss for crowd count estimation with point supervision,''
\newblock in {\em 2019 IEEE/CVF International Conference on Computer Vision
  (ICCV)}, 2019, pp. 6141--6150.

\bibitem{VGG}
Karen Simonyan and Andrew Zisserman,
\newblock ``Very deep convolutional networks for large-scale image
  recognition,''
\newblock in {\em International Conference on Learning Representations}, 2015.

\bibitem{SANet}
Xinkun Cao, Zhipeng Wang, Yanyun Zhao, and Fei Su,
\newblock ``Scale aggregation network for accurate and efficient crowd
  counting,''
\newblock in {\em Computer Vision -- ECCV 2018}, Vittorio Ferrari, Martial
  Hebert, Cristian Sminchisescu, and Yair Weiss, Eds., Cham, 2018, pp.
  757--773, Springer International Publishing.

\bibitem{M-SFANet}
Pongpisit Thanasutives, Ken-ichi Fukui, Masayuki Numao, and Boonserm
  Kijsirikul,
\newblock ``Encoder-decoder based convolutional neural networks with
  multi-scale-aware modules for crowd counting,''
\newblock in {\em 2020 25th International Conference on Pattern Recognition
  (ICPR)}. IEEE, 2021, pp. 2382--2389.

\bibitem{SMANet}
Mingjie Wang, Hao Cai, Jun Zhou, and Minglun Gong,
\newblock ``Stochastic multi-scale aggregation network for crowd counting,''
\newblock in {\em ICASSP 2020 - 2020 IEEE International Conference on
  Acoustics, Speech and Signal Processing (ICASSP)}, 2020, pp. 2008--2012.

\bibitem{983420}
Sheng-Fuu Lin, Jaw-Yeh Chen, and Hung-Xin Chao,
\newblock ``Estimation of number of people in crowded scenes using perspective
  transformation,''
\newblock {\em IEEE Transactions on Systems, Man, and Cybernetics - Part A:
  Systems and Humans}, vol. 31, no. 6, pp. 645--654, 2001.

\bibitem{4761705}
Min Li, Zhaoxiang Zhang, Kaiqi Huang, and Tieniu Tan,
\newblock ``Estimating the number of people in crowded scenes by mid based
  foreground segmentation and head-shoulder detection,''
\newblock in {\em 2008 19th International Conference on Pattern Recognition},
  2008, pp. 1--4.

\bibitem{5206621}
Weina Ge and Robert~T. Collins,
\newblock ``Marked point processes for crowd counting,''
\newblock in {\em 2009 IEEE Conference on Computer Vision and Pattern
  Recognition}, 2009, pp. 2913--2920.

\bibitem{chen2012feature}
Ke~Chen, Chen~Change Loy, Shaogang Gong, and Tony Xiang,
\newblock ``Feature mining for localised crowd counting,''
\newblock in {\em British Machine Vision Conference, {BMVC} 2012, Surrey, UK,
  September 3-7, 2012}, Richard Bowden, John~P. Collomosse, and Krystian
  Mikolajczyk, Eds. 2012, pp. 1--11, {BMVA} Press.

\bibitem{6619163}
Ke~Chen, Shaogang Gong, Tao Xiang, and Chen~Change Loy,
\newblock ``Cumulative attribute space for age and crowd density estimation,''
\newblock in {\em 2013 IEEE Conference on Computer Vision and Pattern
  Recognition}, 2013, pp. 2467--2474.

\bibitem{CP-CNN}
Vishwanath~A. Sindagi and Vishal~M. Patel,
\newblock ``Generating high-quality crowd density maps using contextual pyramid
  cnns,''
\newblock in {\em 2017 IEEE International Conference on Computer Vision
  (ICCV)}, 2017, pp. 1879--1888.

\bibitem{Hydra-CNN}
Daniel O{\~{n}}oro-Rubio and Roberto~J. L{\'o}pez-Sastre,
\newblock ``Towards perspective-free object counting with deep learning,''
\newblock in {\em Computer Vision -- ECCV 2016}, Bastian Leibe, Jiri Matas,
  Nicu Sebe, and Max Welling, Eds., Cham, 2016, pp. 615--629, Springer
  International Publishing.

\bibitem{Geometric}
Di~Kang, Debarun Dhar, and Antoni Chan,
\newblock ``Incorporating side information by adaptive convolution,''
\newblock in {\em Advances in Neural Information Processing Systems}, I.~Guyon,
  U.~V. Luxburg, S.~Bengio, H.~Wallach, R.~Fergus, S.~Vishwanathan, and
  R.~Garnett, Eds. 2017, vol.~30, Curran Associates, Inc.

\bibitem{SpatialPyramidPooling}
Kaiming He, Xiangyu Zhang, Shaoqing Ren, and Jian Sun,
\newblock ``Spatial pyramid pooling in deep convolutional networks for visual
  recognition,''
\newblock {\em IEEE Transactions on Pattern Analysis and Machine Intelligence},
  vol. 37, no. 9, pp. 1904--1916, 2015.

\bibitem{DMCount}
Boyu Wang, Huidong Liu, Dimitris Samaras, and Minh~Hoai Nguyen,
\newblock ``Distribution matching for crowd counting,''
\newblock {\em Advances in Neural Information Processing Systems}, vol. 33, pp.
  1595--1607, 2020.

\bibitem{Adam}
Diederik~P. Kingma and Jimmy Ba,
\newblock ``Adam: {A} method for stochastic optimization,''
\newblock in {\em 3rd International Conference on Learning Representations,
  {ICLR} 2015, San Diego, CA, USA, May 7-9, 2015, Conference Track
  Proceedings}, Yoshua Bengio and Yann LeCun, Eds., 2015.

\bibitem{PyTorch}
Adam Paszke, Sam Gross, Francisco Massa, Adam Lerer, James Bradbury, Gregory
  Chanan, Trevor Killeen, Zeming Lin, Natalia Gimelshein, Luca Antiga, Alban
  Desmaison, Andreas Kopf, Edward Yang, Zachary DeVito, Martin Raison, Alykhan
  Tejani, Sasank Chilamkurthy, Benoit Steiner, Lu~Fang, Junjie Bai, and Soumith
  Chintala,
\newblock ``Pytorch: An imperative style, high-performance deep learning
  library,''
\newblock in {\em Advances in Neural Information Processing Systems 32},
  H.~Wallach, H.~Larochelle, A.~Beygelzimer, F.~d\textquotesingle
  Alch\'{e}-Buc, E.~Fox, and R.~Garnett, Eds., pp. 8024--8035. Curran
  Associates, Inc., 2019.

\end{thebibliography}
\end{document}